\pdfoutput=1
\documentclass[11pt]{article}

\usepackage[final]{acl}

\usepackage{times}
\usepackage{latexsym}

\usepackage[T1]{fontenc}
\usepackage[utf8]{inputenc}

\usepackage{microtype}

\usepackage{inconsolata}

\usepackage{graphicx}
\usepackage{caption}
\usepackage{subcaption}
\usepackage{tabularray}
\usepackage{enumitem}
\usepackage{mathtools}
\usepackage{amssymb}
\usepackage{amsthm}
\usepackage{amsmath}
\usepackage{multirow}
\usepackage{longtable}
\usepackage{booktabs}

\usepackage{todonotes}

\usepackage[ruled,lined,linesnumbered]{algorithm2e}
\def\BibTeX{{\rm B\kern-.05em{\sc i\kern-.025em b}\kern-.08em
    T\kern-.1667em\lower.7ex\hbox{E}\kern-.125emX}}

\newcommand{\squishlist}{ 
   \begin{list}{$\bullet$}
    { \setlength{\itemsep}{0pt}      \setlength{\parsep}{3pt} 
      \setlength{\topsep}{3pt}       \setlength{\partopsep}{0pt}
      \setlength{\leftmargin}{1.5em} \setlength{\labelwidth}{1em}
      \setlength{\labelsep}{0.5em} } }
\newcommand{\squishend}{
    \end{list}  } 

\title{ViClaim: A Multilingual Multilabel Dataset for Automatic Claim Detection in Videos}

\author{
Patrick Giedemann\textsuperscript{1}, Pius von Däniken\textsuperscript{1}, Jan Deriu\textsuperscript{1}, \\ {\bf Alvaro Rodrigo\textsuperscript{2}}, {\bf Anselmo Peñas\textsuperscript{2}}, 
{\bf Mark Cieliebak\textsuperscript{1}} \\
\textsuperscript{1}Zurich University of Applied Sciences, Winterthur \\
\textsuperscript{2} UNED NLP \& IR Group, Spain \\
gied@zhaw.ch
}

\begin{document}
\maketitle

\begin{abstract}
The growing influence of video content as a medium for communication and misinformation underscores the urgent need for effective tools to analyze claims in multilingual and multi-topic settings. Existing efforts in misinformation detection largely focus on written text, leaving a significant gap in addressing the complexity of spoken text in video transcripts. We introduce ViClaim, a dataset of 1,798 annotated video transcripts across three languages (English, German, Spanish) and six topics. Each sentence in the transcripts is labeled with three claim-related categories: fact-check-worthy, fact-non-check-worthy, or opinion. We developed a custom annotation tool to facilitate the highly complex annotation process. Experiments with state-of-the-art multilingual language models demonstrate strong performance in cross-validation (macro F1 up to 0.899) but reveal challenges in generalization to unseen topics, particularly for distinct domains. Our findings highlight the complexity of claim detection in video transcripts. ViClaim offers a robust foundation for advancing misinformation detection in video-based communication, addressing a critical gap in multimodal analysis.

\end{abstract}

\section{Introduction}
Video content is increasing in popularity worldwide. Alone in the US, the average adult spends around 47 minutes on YouTube and 55 Minutes on TikTok~\footnote{\url{https://www.statista.com/statistics/1359403/us-time-spent-per-day-netflix-tiktok-youtube}}.  Especially short-format videos (i.e., videos of at most 90 seconds) are gaining in popularity~\footnote{\url{https://www.yaguara.co/short-form-video-statistics/}}. While platforms like YouTube serve educational and informational purposes~\cite{srinivasacharlu2020ytedu,wafa2024correlation}, they are also used to disseminate narratives aimed at influencing viewer's opinions. Despite the recognized necessity of extending misinformation detection technologies to video modalities~\cite{dasanmartin2021suvery_propaganda}, research efforts remain limited, with a predominant focus on analyzing textual data from platforms like X (formerly Twitter)~\cite{arslan2020check,alam2023checkthat}. Furthermore, most existing research on video-based misinformation primarily targets visual features, such as detecting manipulated content~\cite{olga2018fake_video,palod2019misleading,bu2023combating-misinfo-videos}, leaving the semantic analysis of video transcripts largely unexplored, although features found in transcripts of spoken language differ from written language in structure and delivery~\cite{dingemanse-liesenfeld-2022-text}. Studies that do analyze transcripts tend to treat them as a global, binary classification task, aiming to determine whether a video contains misinformation as a whole~\cite{hou2019towards,hussein2020yt-audit,papadamou2022yt-pesudoscience,christodoulou2023identifying}. However, misinformation detection is a multilayered task, with claim detection as the first step~\cite{PANCHENDRARAJAN2024}. 

In this work, we present the first step towards misinformation detection in videos by introducing  ViClaim, a novel dataset where transcripts of YouTube Short videos are annotated using a custom annotation tool on a sentence level with the claim taxonomy introduced in~\cite{PANCHENDRARAJAN2024}.  Each sentence is annotated to indicate whether it contains a claim, an opinion, or both, and if a claim is present, whether it is check-worthy. Since sentences may fall into multiple categories simultaneously, the task is framed as a multi-label classification problem. The dataset encompasses 1,798 videos in three languages (English, German, and Spanish) across six topics—five political and one entertainment—allowing for investigating domain transfer capabilities. Each sentence was annotated by four independent annotators, resulting in a total of 17,116 annotated sentences. Thus, ViClaim offers a rich database for future research on multimodal misinformation detection. As a first step to showcase the utility of  ViClaim, we trained several baseline models on the transcripts (leaving multimodal models for future work). Our best-performing model achieved an F1 score of 89.9 for checkworthiness classification, 77.6 for detecting non-check-worthy claims, and 83.6 for opinion detection. These results demonstrate both the effectiveness of the dataset and the challenges posed by the nuanced task of analyzing short-form video transcripts. 
To summarize, our contributions are as follows: (i) the ViClaim dataset, a multilingual, multi-topic resource for claim detection in video transcripts; (ii) a custom annotation tool and comprehensive guidelines; (iii) baseline models to showcase the potential of ViClaim. %, and (iv) the open-sourcing of all resources, including the annotated transcripts with the raw annotations, video IDs, and experimental code.

\paragraph{Release.} We release the corpus in form of the video IDs used, the time stamps of the annotated sentences, the corresponding labels, and the code pipeline to reconstruct the corpus, which can be found under the following GitHub repository~\footnote{\url{https://github.com/pgied/viclaim_stt}}. We release the experimental pipeline and the trained models, which can be found under the following GitHub repository~\footnote{\url{https://github.com/pgied/viclaim_training}}.

\section{Related Work}
Claim detection constitutes the first step in misinformation detection~\cite{PANCHENDRARAJAN2024} where we mainly refer to claims that, due to their content, have a significant impact on public opinion or pose a risk of being widely disseminated due to their controversial nature \cite{alam2023checkthat}. There has been vast research on claim detection, mainly on texts \cite{zhang2020overview}, where most systems receive the exact statement to be checked. These claims are usually sentences extracted from a document or social media posts (e.g., a Tweet) \cite{barron2020checkthat}. On the other hand, other datasets aimed at detecting the exact span containing the claim in tweets, which is a more challenging task, but it allows systems to retrieve more accurate evidence \cite{sundriyal2022empowering}. The available datasets commonly focused on a single and controversial topic like environment \cite{stammbach2023environmental-claim}, politics \cite{Dutta2023} or COVID-19 \cite{Faramarzi2023}. Some datasets contain claims about several topics like elections or COVID-19 \cite{kazemi-etal-2021-claim}. The most common language for these datasets is English. Still, there have also been several efforts for creating datasets in multiple languages, like the dataset from the NLP4IF 2021 shared task, with tweets in English, Arabic, and Bulgarian \cite{shaar2021findings}, extended in the CLEF2022 CheckThat! lab with Dutch tweets \cite{alam2021checkthat-main}.

Although most of the work on claim detection has been devoted to text content, there have also been some efforts towards detecting claims in multimedia content. These works consider that multimedia content in social networks is an effective way for spreading misinformation compared with textual content \cite{jin2017multimodal,dhawan2022game}. One of the first works in the multimedia setting extended previous text-based collections with images to evaluate multimodal detection approaches  \cite{Cheema21}. However, the labels were only based on the textual content. This is why \citet{cheema-etal-2022-mm} created the MM-Claims dataset, which includes tweets and corresponding images and where the annotations are based on both text and images. The CheckThat! Lab 23 followed their methodology  \cite{alam2023checkthat}, where the organizers show the tweet's image alongside the tweet itself, and the image could be a piece of evidence or contain a text containing a claim. The inclusion of multimodal data yielded better scores~\cite{von2023zhaw}.

On the other hand, to the best of our knowledge, there is almost no work on claim detection using videos. The most relevant work is the ClaimBuster dataset \cite{arslan2020check}, which provides 23,533 statements extracted from transcripts of US general election presidential debates and annotated by humans for check-worthiness. This work is somewhat similar to our dataset, although it only provides the statements to be checked without context, in English, and without the original audio or video.  
Most of the previous works related to videos have focused on the task of detecting whether they contain misinformation~\cite{hou2019towards,hussein2020yt-audit,papadamou2022yt-pesudoscience,christodoulou2023identifying}. \citet{Micallef_2022} found that traditional approaches focused on textual content and missed post-video pairs that contain misinformation. They improved detection results when considering features from platforms containing the linked videos. \citet{choi2022fakeYT} developed a deep learning model that integrates different modalities for detecting if full YouTube videos were fake or real. \citet{palod2019misleading} developed VAVD, a video dataset with fake and non-fake annotations. Besides, they obtained promising results by relying only on user comments. \citet{olga2018fake_video} created the InVID Fake Video Corpus, with annotations for the full videos. 

%Nevertheless, all these works give the same importance to all the claims without including additional information or a taxonomy about the type of language used in the statement containing the claim.

\section{Data Collection}
This section describes the data collection efforts for claim detection in video. The final corpus contains 1798 videos in short format. Each video was transcribed and split up into sentences, which resulted in a total of 17116 sentences. Each sentence has been annotated by four different annotators. The dataset spans three languages: English, German, and Spanish~\footnote{The language selection is based on the languages spoken by the authors of this work, which allowed us to monitor the annotation process.}. We first describe the annotation tool together with the task; then, we describe the label set, the topics of the videos, the video selection, and the annotator agreement computation. The data collection ran from 12. April to 3. June 2024 and from 28. October to 25. November 2024.

\subsection{Annotation Tool}
\begin{figure*}[!t]
\centering
\small
     \includegraphics[width=0.8\textwidth]{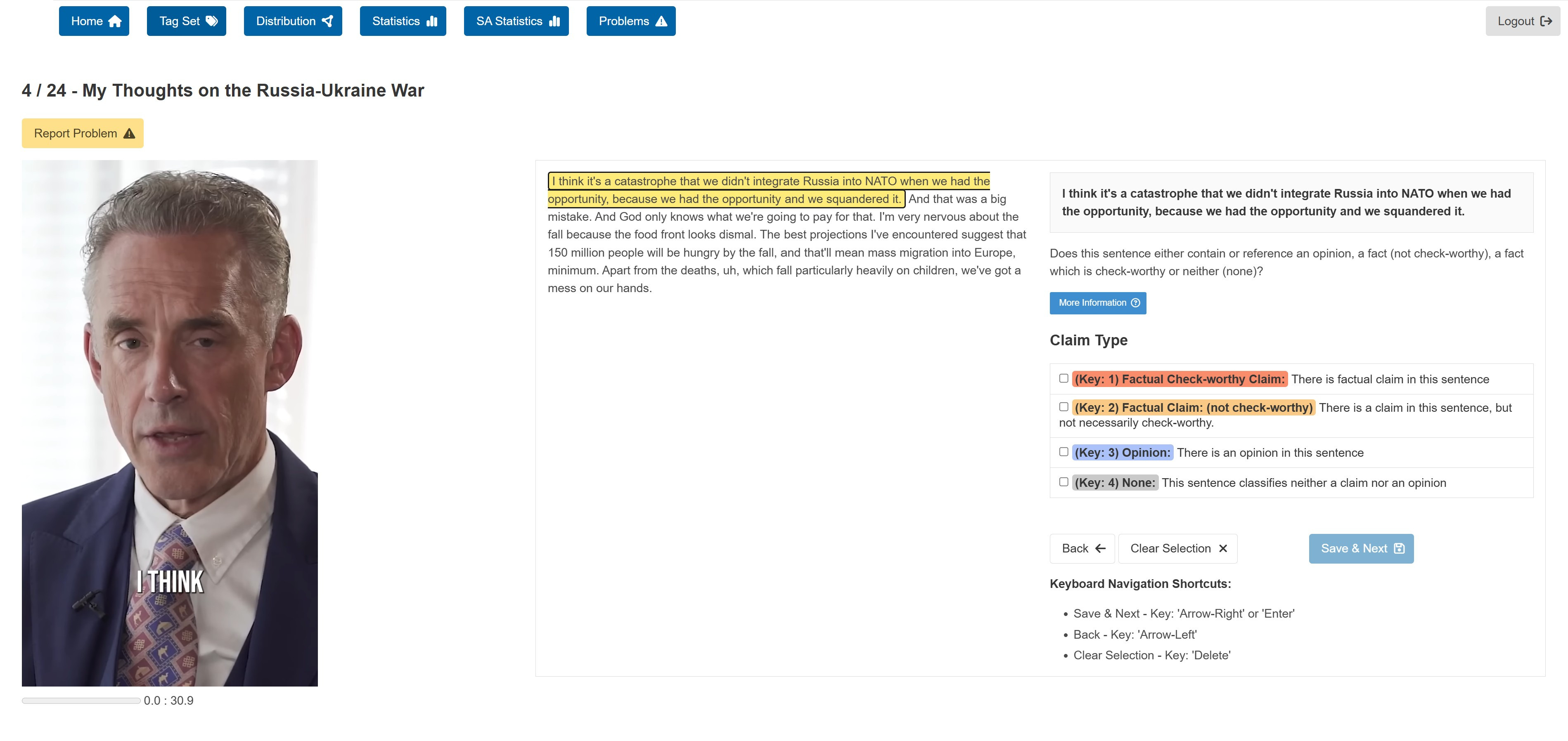}
\caption{Annotation Tool. The user is shown a short video and the transcript, which is split by sentence, and then they annotate each sentence with one or more of the four tags.}
\label{fig:annotation_tool}
\end{figure*}

The annotation tool (see Figure~\ref{fig:annotation_tool}) displays the video alongside its transcript. Annotators are tasked with watching the video and labeling each sentence using the provided tagset: fact-check-worthy, fact-non-check-worthy, opinion, or none (cf. \ref{subsec:tag_set_description}). Following the approach of~\cite{arslan2020check}, we collect annotations at the sentence level, as allowing annotators to define their own spans often introduces excessive noise.

While we recognize that sentences may contain multiple claims (some check-worthy, some not) or combine opinions and claims and that some may span multiple sentences (see Table~\ref{tab:example_sentences}), we address these complexities by employing a multi-label annotation approach. This ensures comprehensive coverage without compromising consistency. Importantly, we tested allowing annotators to define custom spans, which led to significant disagreement and unusable data due to varying interpretations of span boundaries and overlaps. As a result, sentence-level annotation was chosen as a more consistent and reliable alternative~\footnote{Previous experiments with custom spans resulted in Krippendorff’s $\alpha$ scores below 0.2, making the approach impractical.}.

The transcripts were created using AssemblyAI~\footnote{\url{https://www.assemblyai.com}. We opted for a paid solution as it provides diarization out of the box.}, and the sentence segmentation was performed using SpaCy~\footnote{\url{https://spacy.io}}.

\subsection{Tag-Set Description}
\label{subsec:tag_set_description} % Label for referencing
Our tagset is based on the taxonomy introduced in~\cite{PANCHENDRARAJAN2024}, which categorizes claims into factual and opinions, and factual are further divided into check-worthy facts and non-check-worthy (either due to being non-verifiable or not check-worthy). Thus, we introduce the three labels: Fact Check-worthy (FCW), Fact Non-Check-worthy (FNC), and Opinion (OPN). Note that in contrast to~\cite{arslan2020check}, which only differentiates between Check-worthy Factual Statements and Uncheckable Factual Statements, we explicitly label opinions to capture subjective elements as well as broader contextual and persuasive aspects that are essential for understanding misinformation~\cite{goldberg2021opinion}. Additionally, we adopt a multi-label annotation approach, recognizing that sentences often encompass multiple relevant categories. 

%Our tag set builds upon the framework of the ClaimBuster dataset~\cite{arslan2020check}, which employs three labels: Check-worthy Factual Statement (CFS), Uncheckable Factual Statement (UFS), and Non-Factual Statement (NFS). These correspond to our labels: Fact Check-worthy (FCW), Fact Non-Check-worthy (FNC), and None, respectively. We enhance this framework by introducing an additional tag, Opinion (OPN), to capture subjective elements such as opinions, beliefs, and speculations. 
%This extension provides a more nuanced and comprehensive basis for downstream analysis, enabling richer insights into the data.

\noindent{\bf Fact that is check-worthy (FCW).} Sentences containing factual claims of public interest that are verifiable and relevant for fact-checking, commonly sought by journalists.

\noindent{\bf Fact that is not check-worthy (FNC).} Factual claims that are either unverifiable or lack public interest, such as personal experiences or jokes. 

\noindent{\bf Opinion (OPN).} This tag encompasses subjective sentences, including opinions, beliefs, accusations, speculations, predictions, and emotional expressions. 

%It was introduced to capture broader contextual and persuasive elements that are essential for understanding misinformation. As highlighted in \cite{goldberg2021opinion}, "The differentiation between factual information and opinions is central to journalistic norms," and as further emphasized in \cite{smith2018opinion}, "Distinguishing facts from opinions may significantly improve subsequent analytics tasks." 

\noindent{\bf None.} Sentences that do not fit the above categories, such as commands, insults, casual expressions, or threats.

For examples of each tag and their combinations, see Table~\ref{tab:example_sentences}.

\begin{table*}[t!] % Use table* for full-width table
    \centering
    \small
    \resizebox{.95\textwidth}{!}{%
    \setlength{\tabcolsep}{6pt} % Adjust cell padding (horizontal spacing)
\renewcommand{\arraystretch}{1.4} % Adjust row height (vertical spacing)

\begin{tabular}{ p{0.4\textwidth} p{0.1\textwidth} p{0.5\textwidth} }
    \hline
    \multicolumn{1}{c}{\textbf{Sentence}} & 
    \multicolumn{1}{c}{\textbf{Labels}} & 
    \multicolumn{1}{c}{\textbf{Explanation}} \\ \hline
    10,000 immigrants arrive daily. & FCW & Clearly checkable number and relevant to society. \\ \hline
    And that's what people over and over told me, that, of course. & FNC & This claim cannot be verified and is therefor not check-worthy. \\ \hline
    The company will probably go bankrupt within a year. & OPN & This is clearly an opinionated speculation.\\ \hline
    Don’t ever use the word smart with me. & NONE & This is a threat and is neither an opinion nor a fact. \\ \hline
    He's first to admit that, uh, and he's pretty profane at times when he's fired up about something, and certainly he is about Donald Trump. & FNC, OPN & This sentence contains the labels FNC and OPN, since the first part is not possible to check, there is not verifiable information and the middle part is a subjective view (opinion). \\ \hline
    These politicians will lie to your face and make millions while normal Americans pay the price. & FCW, OPN & This sentence contains the labels FCW and OPN, since the first part is a subjective view and the second part is verifiable and has a public relevancy. \\ \hline
    I don't have them in front of me, but we're going to, if Pence becomes a candidate, we will look at that in more detail. & FNC, OPN & This sentence contains the labels FNC and OPN, since the first part is a self description and not not really publicly relevant and the second part is a future prediction which is not yet a fact but rather an intention. \\ \hline
\end{tabular}
    }
    \caption{Examples of sentences within the transcripts along with their ground-truth labels, and explanations }
    \label{tab:example_sentences} % Assign a label
\end{table*}

\subsection{Topic Selection}
We selected five highly relevant socio-political topics during the video selection process conducted in May 2024 and November 2024. These topics were chosen based on their international relevance and widespread public interest, although it is important to acknowledge that the selected videos reflect a bias toward the Western world. The aim was to ensure coverage of topics that would generate content in multiple languages, enabling a diverse and multilingual dataset. Below, we describe the selected topics:
\begin{itemize}[noitemsep]
    \item US Elections 2024. This topic includes videos focusing mostly on the two leading candidates, Donald Trump and Joseph Biden~\footnote{Since most of the video selection was conducted in May 2024, not many videos account for the candidacy of Kamala Harris.}. The focus on these candidates allows recognition beyond the United States, as we included videos in both German and Spanish in addition to English.
    \item War in Ukraine. This topic covers the conflict between Ukraine and Russia and has generated videos in multiple languages, making it suitable for our multilingual dataset.
    \item Migration. Videos on this topic discuss various aspects of migration, providing content in multiple target languages.
    \item European Union. This topic includes perspectives on the EU, particularly around the European Parliament elections in June 2024, with videos available in English, German, and Spanish.
    \item General view about the USA. This category includes perspectives on US society from both internal and external viewpoints, addressing sociopolitical dynamics.
\end{itemize}
To evaluate the claim detection approach in a domain transfer setting, we included an unrelated topic: League of Legends~\footnote{\url{https://www.leagueoflegends.com/de-de/}}, a globally popular online multiplayer video game. This topic was selected to evaluate the dataset’s applicability to different domains and support experiments in out-of-domain generalization.

\subsection{Video Selection}
Creating video annotations requires significant effort and resources, as it depends on identifying videos that explicitly contain claims. Preliminary tests revealed that semi-automated approaches, such as keyword-based searches, often returned videos irrelevant to the target topics or devoid of claims. We opted for a manual video selection process due to the inherent challenges of using YouTube's recommender system, as highlighted by~\cite{chandio2024audit} and our experience. Their findings show that factors like strong recency bias, the choice of seed videos, and the depth of exploration significantly influence the diversity and characteristics of recommended content. These complexities make automated selection prone to biases that are difficult to control, potentially limiting its effectiveness in curating a representative dataset.

To address these challenges, two trained researchers fluent in all three target languages manually searched for short-form videos (no longer than 90 seconds) across six predefined topics. New YouTube accounts were created to minimize recency and personalization biases. They used structured keyword searches tailored to each topic and refined keywords iteratively to ensure diversity. For example, keywords for the US Elections included "US elections 2024 Biden," "Trump policies," and "election debates." A mix of popular and niche videos was selected to capture diverse content and viewpoints, particularly for contentious topics like Migration and the US Elections.

The manual selection yielded 1798 short-form videos across six topics and three languages (approx. 100 videos per topic-language pair, cf. Appendix~\ref{app:dataset_distribution_overview}).

\subsection{Annotation Management}
\label{subsec:annotation_management}
Due to the complexity of the annotation task, we opted against crowdsourcing and contracted twelve annotators (6 female, 6 male, ages 18–34). Our annotators included seven native German speakers (two also proficient in Spanish), two native Spanish speakers, and three bilingual English/German or Spanish/German speakers. All but two were students, and all were highly proficient in English, German, or Spanish. Each annotator was assigned 600 videos, grouped as follows:

\begin{itemize}[noitemsep]
    \item Group 1: Four annotators annotated 300 videos in English and 300 in German.
    \item Group 2: Four annotators annotated 300 videos in English and 300 in German (distinct from Group 1).
    \item Group 3: Four annotators annotated 600 videos in Spanish.
\end{itemize}

We compensated the annotators with 500 euros for a clean completion. This corresponds to an hourly salary of 25 Euros. Following~\citet{bender-friedman-2018-data}, we let the annotators complete a questionnaire to rate our task and collect information about their stances on the six topics (a properly anonymized version is available in Appendix~\ref{sec:exit_form}).

\paragraph{Participant Training.}
The training process contained guidelines, examples, workshops, and ongoing support for annotators throughout the annotation process.

First, we provided the annotators with guidelines to instruct them on the usage of the annotation tool, the tagset explanation (cf. \ref{subsec:tag_set_description}), and examples, including edge cases and ambiguous examples. The set of examples was dynamically extended when new edge cases were discovered. We began the training of the annotators with a kickoff workshop to introduce annotators to the project, annotation tool, and task requirements. Annotators were asked to complete 10–20 annotations within the first two days. This allowed us to monitor their understanding and provide targeted feedback based on their early performance. The second workshop occurred after the first 100–200 annotations. Annotations with low agreement were reviewed and discussed in detail to identify patterns of misunderstanding. We maintained an open line of communication, promptly addressing any emerging questions or challenges. This proactive approach ensured annotators remained confident and consistent in their work.

\subsection{Quality Control}  
Since the task is inherently ambiguous, we established the following process to ensure annotation quality. Similar to~\cite{arslan2020check}, two authors collaboratively created a gold standard by annotating 30 videos per language (5 per topic), which resulted in 833 gold sentences. This collaborative effort ensured a consistent interpretation of claims across topics and languages. We continuously tracked annotator agreement with the gold standard, intervening directly (by reaching out to the annotators) when deviations occurred to maintain alignment and improve annotator understanding.

To further ensure the reliability of the annotations, we monitored inter-annotator agreement (Krippendorff's $\alpha$) and pairwise agreement using Jaccard similarity scores for the multi-label annotations over time. This continuous tracking allowed us to detect and address inconsistencies early in the process, thereby improving the overall quality of the dataset.

\paragraph{Annotator Agreement.} 
We evaluated annotator agreement using two metrics. First, we calculated Cohen's $\kappa$ between individual annotators and the gold standard annotations. Second, we computed Krippendorff's $\alpha$ to measure agreement across all annotators. Table~\ref{tab:agreements} presents the agreement scores for each of the three groups.

For Cohen's $\kappa$, we report the average and maximum values across the four annotators in each group. Agreement levels varied by label:
\begin{itemize}[noitemsep]
    \item FCW: Agreement is moderate on average ($0.4 - 0.59$), with Group 3 achieving substantial agreement ($0.64 - 0.76$).
    \item FNC: Agreement is lower, with fair levels on average ($0.35 - 0.52$), though Group 3 achieved a maximum of $0.69$.
    \item OPN: Agreement is moderate ($0.47 - 0.58$ on average), with Group 3 again demonstrating higher consistency ($0.58 - 0.66$).
    \item None: This label consistently achieved the highest agreement, ranging from $0.57 - 0.85$, with averages close to substantial levels.
\end{itemize}

For Krippendorff's $\alpha$, scores ranged from $0.415$ (Group 2) to $0.522$ (Group 3), reflecting moderate agreement overall. Although these agreement levels are not particularly high, they are consistent with the inherent difficulty and subjectivity of the task. A closer analysis revealed that most ambiguity stems from the scenario where multiple classes are appropriate, and two annotators chose non-overlapping subsets of these appropriate classes (see Appendix~\ref{app:ambiguous_sents} for examples of such ambiguous sentences). Thus, the disagreement often did not stem from poor annotator behavior but from the inherent ambiguity. Generally, it has been shown that differentiating between factual statements and opinions is a difficult task~\cite{goldberg2021opinion}. 

In our case, at least one annotator in each group consistently demonstrated moderate to substantial agreement with the gold standard annotations, which we used as a basis for label normalization. This ensures that the resulting dataset retains high reliability despite the task's unavoidable ambiguity challenge.

\paragraph{Label Normalization.}  
\label{par:label_norm}
To generate the final labels for the dataset, we leverage the observation that some annotators demonstrate higher agreement with the gold standard annotations. To account for this, we use MACE~\cite{hovy2013learning}, a Bayesian Model used to compute a trust score for each annotator, reflecting their reliability. We normalize these trust scores to sum to 1 across all annotators. %Using this information, we derive two types of labels: hard labels and soft labels.

%\noindent{\bf Hard Label.} The hard label, denoted as $y_{\text{hard}} \in \{0, 1\}^3$, is a binary vector of length three where each position corresponds to one of the three primary classes (FCW, FNC, and OPN, respectively).  If all entries in $y_{\text{hard}} = 0$, the sentence is assigned the "None" label, indicating that no claim was identified. The hard labels are the final label for evaluating the models trained on our data. Since MACE only works for multi-class single-label problems, we first convert the binary annotations from individual annotators into one of eight possible combinations, treating the binary vector as a binary number. The process involves the following steps: (i) let $a_i \in \{0, 1\}^3 $ be the annotation of annotator $i$ for a sample, (ii) convert $a_i$ into a decimal value and then into a one-hot encoded vector $b_i \in \{0, 1\}^8$, (iii) compute the weighted sum of all annotators’ one-hot vectors using their trust scores $w_i$, $b = argmax(\sum_i w_i*b_i)$, finally (iv) Convert the resulting $b$ back into its binary representation, yielding $y_{\text{hard}}$.

\noindent{\bf Soft Label.} Based on the trust scores, we derived soft-labels, which capture the ambiguity of the annotations. The soft label, denoted as $y_{\text{soft}} \in [0, 1]^3$, provides a probabilistic interpretation of the label assignments. Each entry represents the probability of the corresponding label. To create the soft label, we simply compute $y_{\text{soft}} = \sum_i w_i*a_i $, where $w_i$ denotes the normalized trust score of annotator $i$ and $a_i \in \{0,1\}^3$ is the annotation of annotator $i$. This approach ensures that the soft label incorporates annotator trust and provides nuanced information about the likelihood of each label.

\begin{table}[t!]
    \centering
    \small
    \resizebox{.48\textwidth}{!}{%
    \setlength{\tabcolsep}{6pt} % Adjust cell padding (horizontal spacing)
\renewcommand{\arraystretch}{1.2} % Adjust row height (vertical spacing)

\begin{tabular}{c|ccc}
      ~  & Group 1 & Group 2 & Group 3 \\ \hline
     FCW & 0.51 | 0.59 & 0.57 | 0.70 & 0.64 | 0.76 \\
     FNC & 0.38 | 0.52 & 0.35 | 0.48 & 0.42 | 0.69 \\
     OPN & 0.47 | 0.53 & 0.53 | 0.68 & 0.58 | 0.66\\
     None& 0.70 | 0.77 & 0.57 | 0.85 & 0.68 | 0.81\\ \hline
$\alpha$ & 0.442 & 0.415 & 0.522\\
\end{tabular}

    }
    \caption{Agreement scores. Cohen's $\kappa$ between the annotators and the gold annotations, we report the average and maximum overall annotators for each group. The last column shows the Krippendorf $\alpha$-score for the annotator agreement. }
    \label{tab:agreements}
\end{table}

\subsection{Data Overview}
\begin{figure}[!t]
\small
\centering
\includegraphics[width=0.49\textwidth]{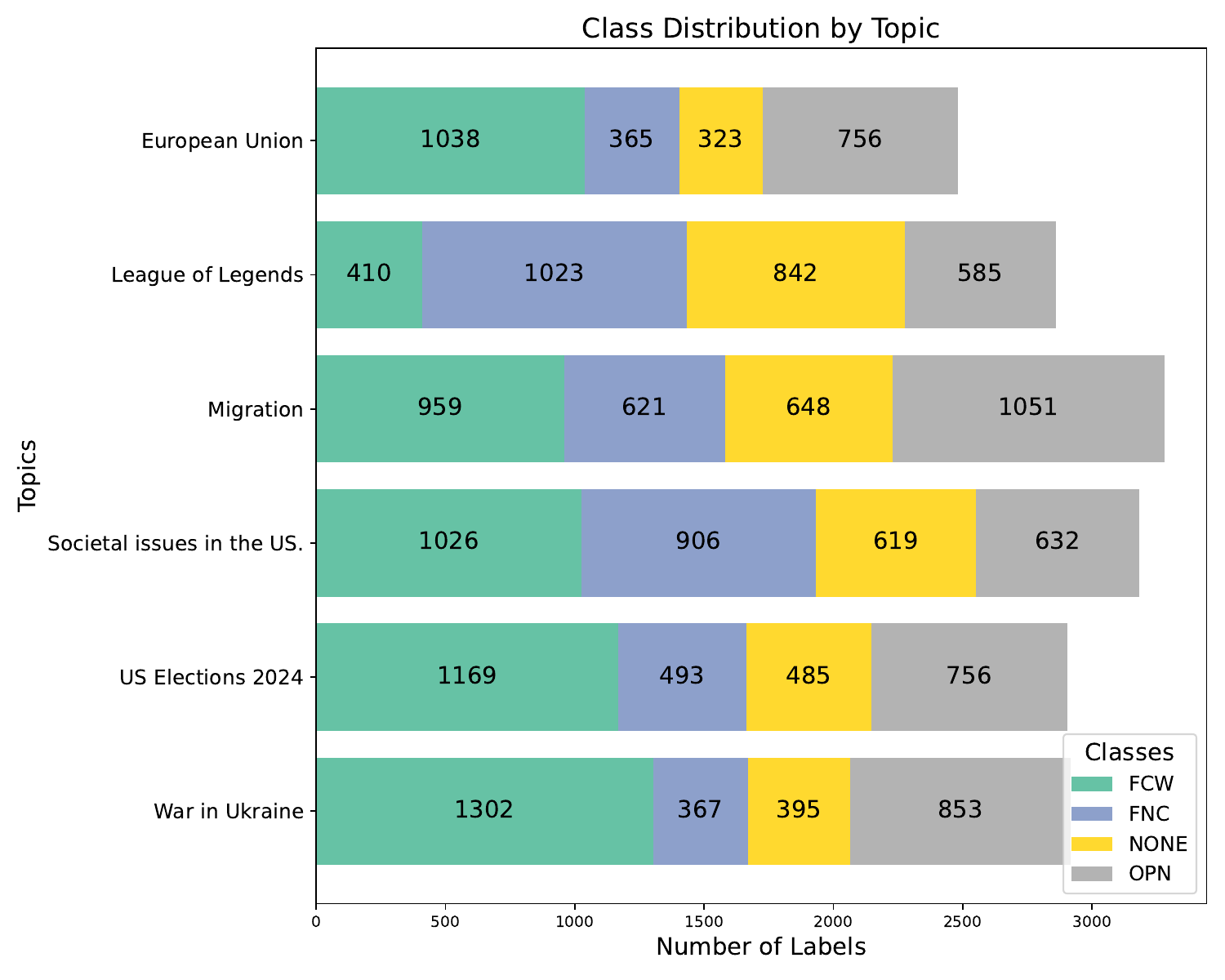}
\caption{For each label, each topic's appearance percentage is depicted. The labels are sorted by their overall frequency. }
\label{fig:tag_distr}
\end{figure}

The resulting dataset contains 1,798 videos, approximately 100 videos per language and topic. In total, there are 17,116 annotated sentences. Figure~\ref{fig:tag_distr} illustrates the distribution of labels across the six topics.

While political topics, such as the US Elections and the War in Ukraine, have many Fact-check-worthy (FCW) samples, the League of Legends topic exhibits a significantly lower prevalence of FCW labels. Instead, it shows a higher prevalence of Fact-Non-check-worthy (FNC) labels. This variation highlights the differences like the topics, with political content being more focused on verifiable claims, while non-political content tends to include more subjective or non-verifiable statements.

\section{Baseline Experiments}
Here, we describe the baseline experiments, for which we fine-tune four different pre-trained multilingual large-language models (LLM). These experiments serve as starting points, leaving more sophisticated and multimodal approaches to future work. 

\subsection{Experimental Setting}

\paragraph{Data.} The models are trained to predict the labels of each sentence in a transcript. For this, let $\mathcal{C} = \{c_i\}_{i=1}^{1798}$ be the set of all clip transcripts. Each transcript is segmented into sentences that are labeled, thus $c_i = \{s_j^{(i)}\}_{j=1}^{n_i}$ where $n_i$ is the number of sentences of transcript $c_i$. Each sentence is a pair of text and label $s_j^{(i)} = (x_j^{(i)}, y_j^{(i)})$, where $x_j^{(i)} = \{t_k\}_{k=1}^{m_{ij}}$ denotes the tokens ($m_{ij}$ number of tokens in sentence j of clip i), and $ y_j^{(i)}$ is the label (cf.~\ref{par:label_norm}). The input to the classifiers is the concatenation of the full clip transcript for the context, with the sentence of the transcript to be classified, i.e., $\mathcal{D} = \{([x_1^{(i)};..;x_{n_i}^{(i)};x_j^{(i)}], y_j^{i})\}$, where $i$ denotes the clip number, $j^{th}$ the sentence within clip $i$ to be classified, and $[;]$ denotes the concatenation of sentence strings. During training, we use the soft labels as in~\cite{fornaciari2021soft} by computing the cross-entropy loss between the system output and the soft labels. 

\paragraph{Models.}
We selected four state-of-the-art LLMs to be fine-tuned; they are all pre-trained on multi-lingual data containing our three languages of interest. 

\begin{itemize}[noitemsep]
    \item {\bf XLM-Robertal-Large (XLM).}~\citet{conneau-etal-2020-unsupervised} pre-trained an 550M parameter encoder-transformer on 100 languages.
    \item {\bf Falcon-7B (F7B).}~\citet{almazrouei2023falcon} pre-trained a 7B decoder-only model on 1.5T tokens of the RefinedWeb corpus~\cite{penedo2023refinedweb}. The main languages that Falcon performs well on are English, German, Spanish, and French, which cover our use case well.
    \item {\bf Mistral-7B (M7B).}~\citet{jiang2023mistral} pre-trained a 7B parameter decoder-only model. The details of the training data used are undisclosed.
    \item {\bf LLama3.2-3B (L3B).}~\citet{grattafiori2024llama} pre-trained a 3B parameter decoder-only model trained on an undisclosed set of 15T tokens of web data. 
\end{itemize}

While we apply regular fine-tuning on \emph{XLM-Roberta-Large}, we use Quantization and Low-Rank Adapters (QLoRA)~\cite{dettmers2024qlora} to fine-tune the three LLMs~\footnote{\url{https://huggingface.co/docs/peft/index}} (the details are in Appendix \ref{app:hyperparams}).

Generally, we run two types of experiments:~\footnote{For all our experiments, we leverage the Huggingface library \url{https://huggingface.co}}

\noindent{\bf CrossValid.} 5-Fold cross-validation, where we stratify on language and topics. We group the clips according to language and topic, and then we first split a 15\% test set. Based on the other 85\%, we apply standard k-fold CV.

\noindent{\bf Leave Topic Out.} Here, we train on data of 5 topics and use the $6^{th}$ topic as a test set to evaluate the transfer capabilities. During training, we employ early stopping on an evaluation set consisting of the 5 topics used for training. Thus, no information on the left-out topic is available during the training. 

\noindent{\bf Early Stopping, Threshold Selection, and Evaluation.} We compute the Area Under the Receiver Operating Characteristic Curve (AUC)~\cite{marcumj1960roc} score for each label after each epoch on the evaluation set and apply the Youden's J statistic~\cite{youden1950stat} to find the optimal decision threshold for each label. We use the average F1 score over the three labels to decide on the early stopping. We then apply the selected threshold to the test set predictions. We report the macro F1 scores of the test set~\footnote{We use the soft version of \url{https://scikit-learn.org/1.5/modules/generated/sklearn.metrics.f1_score.html} using the sample weight argument to input the soft-labels}. 

\subsection{Results}
\begin{table}[t!]
    \centering
    \small
    \resizebox{.5\textwidth}{!}{%
    \setlength{\tabcolsep}{6pt} % Adjust cell padding (horizontal spacing)
\renewcommand{\arraystretch}{1.2} % Adjust row height (vertical spacing)

\begin{tabular}{|l|l|l|l|}
\hline
    & FCW & FCN & OPN \\ \hline
    F7B & $0.889\pm0.003$ & $0.757\pm0.008$ & $0.823\pm0.007$ \\ \hline
    L3B & $0.898\pm0.002$ & $0.772\pm0.008$ & $0.833\pm0.004$ \\ \hline
    M7B & $0.891\pm0.006$ & $0.765\pm0.005$ & $0.829\pm0.008$ \\ \hline
    XLM & $\bf 0.899\pm0.002$ & $\bf 0.776\pm0.007$ & $\bf 0.836\pm0.004$ \\ \hline
\end{tabular}
}
    \caption{\emph{CrossValid} F1-scores for the 4 different models and each label. The overall score is the macro F1 score. The best score in each row is in bold.}
    \label{tab:label_res_cv}
\end{table}

\begin{table}[t!]
    \centering
    \small
    \resizebox{.49\textwidth}{!}{%
    \setlength{\tabcolsep}{6pt} % Adjust cell padding (horizontal spacing)
\renewcommand{\arraystretch}{1.2} % Adjust row height (vertical spacing)

\begin{tabular}{|l||l|l|l|l|}
\hline
    Left-Out-Topic:            & F7B  & L3B  & M7B  & XLM \\ \hline
    European Union      & 0.765 & 0.787 & 0.777 & 0.793 \\ \hline
    League of Legends   & \it 0.690 & \it0.707 & \it0.720 &\it 0.721 \\ \hline
    Migration           & 0.726 & 0.768 & 0.766 & 0.763 \\ \hline
    Views about US  soc.      & \bf0.784 & \bf 0.796 & \bf 0.790 & \bf 0.798 \\ \hline
    US Elections        & 0.740 & 0.763 & 0.738 & 0.768 \\ \hline
    War in Ukraine      & 0.752 & 0.766 & 0.770 & 0.753 \\ \hline
\end{tabular}        % 
}
    \caption{\emph{Leave Topic Out} experiments for the label detection. We report the macro F1 score for each Model and left-out topic. The best score in each column is in bold, and the worst score is in italics.}
    \label{tab:label_res_topic}
\end{table}

\noindent{\bf CrossValid Results} Table~\ref{tab:label_res_cv} presents the F1 scores achieved by the four models for each label in the Cross-Validation (CV) setting. Overall, we observe that the models perform consistently across the labels, with the FCW label achieving the highest F1 scores, averaging around $0.89$. This is likely due to the high prevalence of this label in the training data.
The OPN label also performs well, with scores ranging from $0.823$ to $0.836$. This indicates that the models can effectively distinguish opinions, likely due to their distinct linguistic patterns. 
On the other hand, the FNC label shows slightly lower performance, with F1 scores ranging from $0.757$ to $0.776$. This reflects the label's relatively lower prevalence and greater ambiguity in the data.
Among the models, XLM-Roberta-Large (XLM) achieves the best overall performance, with the highest scores for FCW (0.899), FNC (0.776) and OPN (0.836). This demonstrates that pre-trained multilingual models fine-tuned with sufficient context can excel in multi-label classification tasks.

\noindent{\bf Leave Topic Out Results} Table~\ref{tab:label_res_topic} presents the overall F1 scores for the Leave-Topic-Out experiments. As expected, performance varies depending on the left-out topic, with models performing better on topics closer to the training distribution.

The General views about US society topics achieve the highest scores across all models, ranging from 0.784 to 0.798, likely due to its broader content and overlap with other training topics. The League of Legends topic consistently shows the lowest scores, ranging from 0.690 to 0.721, reflecting its distinct domain and vocabulary compared to the other topics. Performance is moderate for the Migration and War in Ukraine topics (0.726 to 0.770), indicating that the models can generalize moderately well to topics that share argumentative or factual patterns with the training data. European Union and US Elections show slightly higher variability, with scores ranging from 0.740 to 0.793, suggesting that performance depends on the linguistic and contextual overlap with the training set.

\section{Discussion \& Conclusion}
In this study, we introduced ViClaim, a dataset of 1,798 annotated video transcripts spanning three languages (English, German, Spanish) and six topics, including socio-political and non-political content. The dataset employs a multi-label annotation approach, distinguishing between check-worthy and non-check-worthy facts and opinions. We fine-tuned state-of-the-art multilingual models, achieving strong performance in cross-validation (macro F1 up to 0.896). However, generalization to unseen topics remains challenging, particularly for non-political domains like League of Legends, highlighting the need for more advanced approaches to domain adaptation and contextual understanding in video-based misinformation detection. By releasing ViClaim and establishing strong baselines, we aim to drive progress in multi-modal misinformation detection. This dataset addresses a critical research gap and serves as a foundation for developing tools to combat misinformation in the increasingly dominant medium of video.

% In Camera Ready version. Add Michelle who helped with the video selection, and 8 ppl from hamison that thelped
\section*{Acknowledgments}
This work was supported by the CHIST-ERA HAMiSoN project grant CHIST-ERA-21-OSNEM002, by SNF 20CH21 209672 and AEI PCI2022-135026-2.

\section*{Limitations}

\paragraph{Manual Video Selection.} While manual selection has its limitations, it was the most practical option given the complexity of identifying claim-rich videos and the limitations of automated approaches. Although we relied on structured keyword searches to guide our selection, the process inevitably involved subjective choices by the researchers. This introduces potential biases in the dataset, such as overrepresenting certain perspectives or content types. We acknowledge that our approach was not fully systematic, but we prioritized ensuring that the selected videos aligned with the target topics and contained clear claims. By documenting our process and its limitations, we aim to provide transparency and offer a basis for future, more structured improvements in video selection methodologies.

\paragraph{Agreement Scores.} The annotation process encountered challenges due to the subjective nature of claim detection. Disagreement among annotators was not uncommon, as reflected in moderate Krippendorff’s $\alpha$ scores. Although MACE normalization was used to prioritize labels from more reliable annotators, this approach depends on the assumption that these annotators are consistently accurate across all contexts. Additionally, the multi-label annotation framework adds some complexity, especially for sentences with overlapping claims or mixed content, which should be considered when interpreting and applying the dataset to downstream tasks. We note that our agreement scores show a medium agreement according to the standard interpretation. However, this does not imply that our annotations are of low quality, rather it highlights the ambiguity of our task. Thus, future research can work on modeling the uncertainty inherent in such a complex task.

\paragraph{No multimedia analysis.} Currently, the focus lies on the transcripts only. We have not yet analyzed the video and audio features of our corpus. This is part of future work and includes user comments and other meta-data in the analysis of the videos. 

\paragraph{Topic Diversity} We recognize that the dataset’s topical focus does not encompass the full spectrum of claim-rich domains (e.g., health, climate, or science misinformation). Expanding the dataset to include more diverse topics is a natural next step to improve its generalizability and broaden its applicability to other real-world challenges. However, the current selection reflects our prioritization of multilingual socio-political discourse and domain transfer analysis as key research goals for this release."

\paragraph{Challenge of Dataset} The high scores of the classifiers show that the dataset is not highly challenging. However, we are aware that in our field (machine learning and NLP), there is a tendency to require a dataset to pose a challenge so that one can create shared tasks or benchmarks around them. However, our dataset aims to be useful in downstream tasks regarding misinformation detection. Furthermore, the challenge of the datasets lies in domain transfer, where there is a large gap between the F1 score and the in-domain scenario.

\section*{Ethical Considerations}

\paragraph{Compliance with YouTube Terms and Copyright.} We collected videos exclusively from YouTube and took care to comply with YouTube’s Terms of Service (ToS). Each video had a publicly visible URL at the time of selection and a median view count of 19097. We do not redistribute any copyrighted audio-visual content. Instead, we release only the video IDs and the start-to-end timestamps for each annotated sentence. We also provide code that lets researchers recreate the transcripts from these references. The original video materials remain on YouTube. Researchers wishing to access them must comply with YouTube’s ToS. Our dataset neither hosts nor reproduces the videos themselves, thereby avoiding copyright violations.

\paragraph{Privacy of Speakers and Personal Data Mitigation.} Our dataset does not include any direct personal identifiers. We do not disclose private information (e.g., personal addresses, contact details) nor do we release the audio or video footage. We also employed Named Entity Recognition (NER) checks in English, German, and Spanish to ensure the transcripts do not unintentionally reveal non-public personally identifiable information. In instances where sensitive details might have appeared, we would have masked them. Ultimately, we found that named entities refer only to the speakers themselves (who voluntarily published their videos), public figures, or fictional characters (e.g., from “League of Legends”). Therefore, no non-public personal data is disclosed.

\paragraph{Informed Consent.} Since the nature of the task is to watch videos with partially extreme views, we informed the annotators about this and gave them the option to opt out of the task. In one case, an annotator opted out of the annotation task after the trial run. We compensated the annotator for the 1 hour they spent on the trial run and discarded their annotations. 

%\\paragraph{DELETE: Privacy of Speakers.} Since people appear in the videos, we must %\respect their privacy. Thus, our dataset only releases the transcripts with the %\annotations and does not search for any information about the speakers. Furthermore, as %\the speakers put their videos on YouTube, they consent to be in public. We also made %\sure to use videos that had a considerate view count. The median view count in our %\dataset lies at 19097 views at the time of selection. 

\paragraph{Reputation Damage.} There is a chance that there is reputational damage to the speakers in the videos being included in a dataset about misinformation. However, we only annotate whether they claim something and not whether their claims constitute misinformation. 

\paragraph{Bias of Annotators and Researchers.} The choices of the researchers and annotators influence which topics and claims are researched. Thus, there is an implicit bias towards certain topics. For this, we let the annotators fill out a questionnaire to ask about their stances on the various topics in the videos, which can be used to understand the bias. 

\paragraph{Benefits outweigh the potential harm.} We note that spreading misinformation in videos is a highly relevant and important topic. Thus, the benefits of investigating and developing analysis methods to counteract the spread of misinformation outweigh the potential harm done to the speakers.

%\begin{itemize}
%    \item Video with triggering content. Need to make sure annotators are ok
%    \item PII is anonymized
%    \item privacy concerns for speakers/people in the videos
%    \item reputational harm being included in a dataset about misinformation
%    \item potential influence on freedom of expression
%    \item WEIRD bias in both researchers and annotators, influences what is being researched (topics + videos) as well as the annotation procedure (point out questionnaire again!)
%    \item benefits outweigh potential harms
 %   \item maybe people will have to sign some kind of form to access our data? (might not comply with the hamison open data policy)
    
%\end{itemize}
\bibliography{custom}

\appendix

\section{Hyperparameters}
\label{app:hyperparams}
The hyperparameters were selected based on manual trial and error. The budget for GPU computation was limited, and a grid search approach would have been prohibitively expensive. For all models, we used an 8-bit ADAM optimizer via block-wise quantization~\cite{dettmers20218}. The experiments were run on a GPU cluster with 8 NVIDIA H200 GPUs, and the total experimentation time was approximately 100 GPU hours.

\noindent{\bf XLM-Robertal-Large (XLM).}~\citet{conneau-etal-2020-unsupervised} pre-trained an 550M parameter encoder-transformer on 100 languages. For claim detection, we used a learning rate of $5e-04$, a batch size of 32 with a gradient accumulation of 8. 

\noindent{\bf Falcon-7B (F7B).}~\citet{almazrouei2023falcon} pre-trained a 7B decoder-only model on 1.5T tokens of the RefinedWeb corpus~\cite{penedo2023refinedweb}. The main languages that Falcon performs well on are English, German, Spanish, and French, which cover our use case well. We applied QLorA~\cite{dettmers2024qlora} with rank 16, alpha 32, a dropout of 0.05, and a learning rate of $5e-04$. For claim detection we used a batch size of 64 with a gradient accumulation of 8.

\noindent{\bf Mistral-7B (M7B).}~\citet{jiang2023mistral} pre-trained a 7B parameter decoder-only model. The details of the training data used are undisclosed. We applied QLorA~\cite{dettmers2024qlora} with rank 16, alpha 8 a dropout of 0.05 and a learning rate of $5e-04$. For claim detection we used a batch size of 64 with a gradient accumulation of 8.

\noindent{\bf LLama3.2-3B (L3B).}~\citet{llama3modelcard} pre-trained an 8B parameter decoder-only model trained on an undisclosed set of 15T tokens of web data. We applied QLorA~\cite{dettmers2024qlora} with rank 64, alpha 16, and dropout of 0.1, and a learning rate of $5e-04$. For claim detection we used a batch size of 64 with a gradient accumulation of 8.

\section{Dataset Distribution Overview}
\label{app:dataset_distribution_overview}
Table~\ref{tab:datasetoverview} presents the distribution of videos and annotated sentences across topics and languages. Initially, we distributed an equal number of 600 videos to each annotation group, as outlined in \autoref{subsec:annotation_management}.

Given the substantial volume of 1,800 videos, it was not feasible to manually verify the transcripts generated for each video. Instead, annotators were provided with a feature in the annotation tool to flag issues if they encountered problems with the videos. Some transcripts were improperly generated during the transcription pipeline, leading to significant errors, such as missing most of the spoken text or being transcribed into the wrong language. These issues contributed to the irregular distribution displayed in Table~\ref{tab:datasetoverview}.

To address these challenges, we sought to collect new video data with similar content to replace the problematic entries. However, this effort resulted in a slightly reduced and inconsistently distributed collection of annotated videos compared to the initial dataset plan. Table~\ref{tab:datasetoverview} reflects these adjustments and the final dataset composition.

\begin{table*}[t!] % Use table* for full-width table
    \centering
    \small
    \resizebox{0.75\textwidth}{!}{%
    \begin{tabular}{lcccccc}
        \toprule
        & \multicolumn{2}{c}{\textbf{en}} & \multicolumn{2}{c}{\textbf{de}} & \multicolumn{2}{c}{\textbf{es}} \\
        \cmidrule(r){2-3} \cmidrule(r){4-5} \cmidrule(r){6-7}
        \textbf{Topic} & \textbf{Videos} & \textbf{Sentences} & \textbf{Videos} & \textbf{Sentences} & \textbf{Videos} & \textbf{Sentences} \\
        \midrule
        US Elections 2024 & 97 & 1111 & 107 & 1007 & 103 & 692 \\
        Societal issues in the US & 100 & 1401 & 98 & 1133 & 89 & 575 \\
        War in Ukraine & 95 & 922 & 109 & 1009 & 121 & 915 \\
        Migration & 96 & 1211 & 112 & 1235 & 100 & 714 \\
        League of Legends & 97 & 1075 & 90 & 954 & 90 & 760 \\
        European Union & 100 & 879 & 98 & 790 & 96 & 733 \\
        \midrule
        \textbf{Total} & 585 & 6599 & 614 & 6128 & 599 & 4389 \\
        \midrule
        \textbf{Total Videos} & \multicolumn{6}{c}{1798} \\
        \textbf{Total Sentences} & \multicolumn{6}{c}{17116} \\
        \bottomrule
    \end{tabular}
    }
    \caption{Overview of the dataset distribution of videos in each language and for each topic. }
    \label{tab:datasetoverview} % Assign a label
\end{table*}

\begin{table*}[t!] % Use table* for full-width table
    \centering
    \small
    \resizebox{1.0\textwidth}{!}{%
    \setlength{\tabcolsep}{6pt} % Adjust cell padding (horizontal spacing)
\renewcommand{\arraystretch}{1.4} % Adjust row height (vertical spacing)

\begin{tabular}{ p{0.3\textwidth} p{0.2\textwidth} p{0.5\textwidth} }
    \hline
    \multicolumn{1}{c}{\textbf{Sentence}} & 
    \multicolumn{1}{c}{\textbf{Labels}} & 
    \multicolumn{1}{c}{\textbf{Explanation}} \\ \hline
    Russia and China are states that are confident enough to withstand all kinds of pressure. & FNC or OPN & Even if it looks like a factual claim, this is an opinion per guidelines, as it is speculative, subjective, and lacks verifiable information. \\ \hline
    Brexit has failed. & FNC or OPN & Even if it looks like a factual claim, it is hard to proof and seems very subjective. Britain is still a former member of the EU, not a current one. \\ \hline
    Trump was glitching so badly looking bloated and haggard, sweaty and disoriented, gripping the lectern like his life depended on it. & FCW, FNC or/ and OPN & This is a fact-checkworthy claim, as we can verify how Trump appeared in the video. However, the additional description is highly subjective, qualifying it as an opinion as well. Additionally it is questionable if this is really from public interest how Trump looked on a certain video and could therefore also pass as factual not checkworthy claim.\\ \hline
    Maybe because from the time that we get into school, women are being told not to have children, not to aspire to family. & FCW, FNC or OPN & This statement could fit any of the three labels. While it is conveyed as a factual statement and could potentially be verified against school guidelines (though unlikely), it leans toward being a fact-non-checkworthy claim. However, given its opinionated and subjective nature, it could also be classified as an opinion. \\ \hline
    Well, let's just say that it was a different time when Vandal Jacks was made. & FNC or OPN & This statement is presented as a fact but lacks checkworthiness due to insufficient information. Its speculative nature also makes it suitable to classify as an opinion. \\ \hline
    You underestimated the capability and the fighting spirit of the brave ukrainian people. & FCW, FNC or/ and OPN & This statement is presented as a fact but is speculative, with the second part clearly an opinion. It could be classified as fact-checkworthy, fact-non-checkworthy, or opinion, depending on the source and context. Interviews could also be checked to verify whether Russian generals were initially overly confident. \\ \hline
    You misjudged the international community, who has universally condemned your actions. & FCW, FNC or OPN & This statement could be classified as fact-checkworthy, fact-non-checkworthy, or opinion, given its speculative nature. The classification depends significantly on the context and the identity of the person making the claim. \\ \hline
\end{tabular}
    }
    \caption{Examples of sentences where multiple interpretations of tags are valid due to ambiguity. }
    \label{tab:example_difficult_sentences}
\end{table*}

\section{Ambiguous Sentences}
\label{app:ambiguous_sents}
Table~\ref{tab:example_difficult_sentences} shows a set of ambiguous sentences where multiple labels are applicable. These are examples where the annotators disagreed. In most cases, disagreement in the labels stemmed from such sentences, where one annotator only selected one class. In contrast, another annotator selected the other class, but both classes would have been applicable. Thus, for this reason, we opted to work with soft-labels to cover this kind of ambiguity. 

\section{Exit Form Output}

After completing the annotation process, we distributed an exit form to our 12 annotators to gain a pseudonymized view of their biases and perspectives across the six annotation topics. Each annotator used their unique ID to provide responses anonymously. The exit form included questions about demographics, social status, political views, and topic-specific questions to capture opinions and insights. Annotators were allowed to take a side, remain neutral, or leave certain questions unanswered if they preferred not to disclose their opinions. Most of the annotators were students between the age of 25-34 years and native German speakers. Regarding topics for US Elections, we see that most annotators would rather support Biden but still think there should be a more suitable candidate. For the topic of War in Ukraine, we see that most annotators would side with Ukraine and see Russia as the aggressor. For the Migration topic, annotators' opinions were evenly distributed on whether migrants are responsible for increased criminality. Many were unsure about whether migrants are beneficial for their host countries, with some agreeing that migrants fill economic gaps. For the topic of General views about the US society, it is also the annotator's view that the educational system in the US is not quite as good for the broad society. However, they agree that the US is most likely still the most dominant military power in the world. Annotators seem not to be biased against or for US citizens in general. For the topic of the European Union, most annotators agree that the EU is an organization supporting a better world. For the topic of League of Legends, annotators can not really relate due to not having ever played that game, but they would agree that it is rather addictive. We conclude from the exit form, that the annotations made in the dataset would have a minor bias, as described in the summarized results of the form to the corresponding topics. The great majority of annotators stated, that they would do the annotation task again and even would recommend a friend to participate.

%\section{Annotation Process}
%\label{app:merge_algo}
%Annotators were selected based on their subject of study, age and gender to support a possibly wide range of different opinions due their different background. The annotation process began with a kickoff-meeting, where the aim of the project and the annotation tool was explained. The annotators were instructed to annotate between 10 to 20 videos and not more. After they done so their results were discussed bilaterally to ensure a similar understanding of each label and how to label each sentence. After the bilateral session, annotators were provided with a checklist to further support them and give them ideas and rules on when and in which situation to use which label and continue to annotate the next 100 to 200 videos. After that another bilateral session with each annotator took place.  

\section{Closed Source Models}
In a preliminary step, we evaluated three proprietary systems Grok2~\footnote{\url{https://x.ai/news/grok-2}}, o4-mini~\footnote{\url{https://platform.openai.com/docs/models/o4-mini}} and o3-mini~\footnote{\url{https://platform.openai.com/docs/models/o3-mini}} on our classification task. For this, we prompted each model with the task descriptions from our guidelines and added the examples from the paper to the prompt for in-context learning. Their F1 scores are far below the fine-tuned models listed in the table~\ref{tab:closed_source_models_scores}. The best performing models for each respecting label only achieved an F1 score for FCW: 0.7797, F1 score for FNC: 0.5559, and F1 score for OPN: 0.7789. Thus, we abandoned closed-source models for this task and focused on fine-tuning classification models. 

\begin{table}[t!] % Use table* for full-width table
    \centering
    \small
    \resizebox{0.45\textwidth}{!}{%
    \begin{tabular}{l|c|c|c}
\multicolumn{1}{l}{} & \multicolumn{3}{c}{\textbf{F1}} \\ \hline
\textbf{Model} & \textbf{FCW} & \textbf{FNC} & \textbf{OPN} \\ \hline
o3-mini                                             &\bf 0.7797 & 0.5408 &\bf 0.7789 \\
4o-mini                                        & 0.6498 &\bf 0.5559 & 0.7509 \\
Grok 2 (closed source)                        & 0.6650 & 0.5490 & 0.6300 \\  \hline
\end{tabular}
    }
    \caption{Scores for closed source models on the classification task. }
    \label{tab:closed_source_models_scores}
\end{table}

\section{Comparison of Written vs. Spoken Language}
To showcase that spoken and written language differ, we used the XLM-Roberta-based classifiers provided by~\cite{von2023zhaw}, which are fine-tuned on the CheckThat 2023 Twitter data for claim check-worthiness classification~\cite{alam2023checkthat}. For the in-domain data (i.e., test set consisting of Tweets), it achieved an F1 score of 0.693 (with equilibrated precision and recall); however, on our data, it only achieved an F1 score of 0.32, with precision at 0.72, and recall at 0.2. We also fine-tuned a ModernBERT~\cite{warner2024smarter}, which achieved an F1 score of 0.46, with a precision of 0.54 and a recall of 0.40. This shows that the existing written text datasets are inadequate for our usecase. 

\label{sec:exit_form}
\begin{figure*}[!t]
\small
\centering
\includegraphics[width=0.85\textwidth]{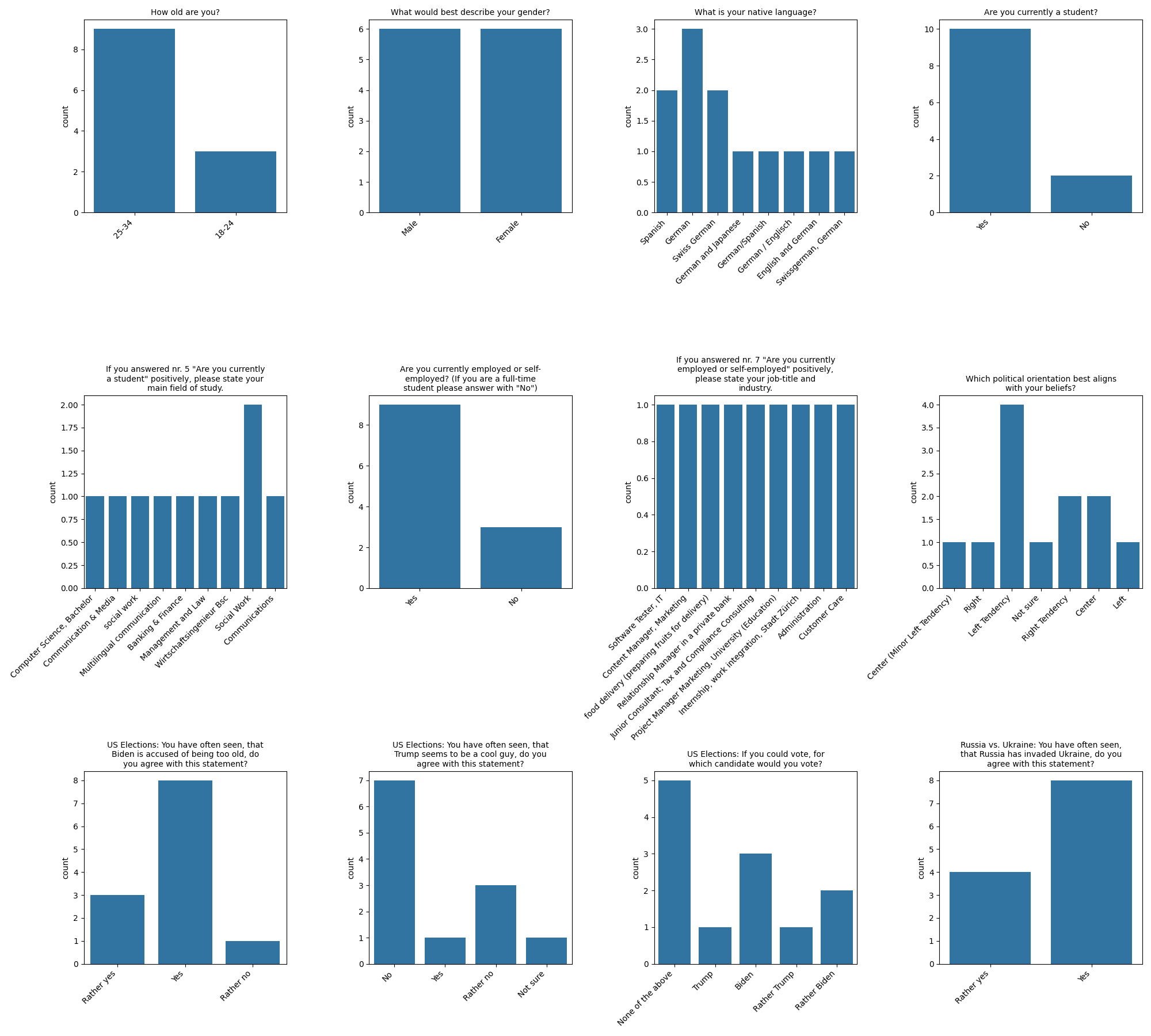}
\caption{Form questions 1 to 12}
\label{fig:form1}
\end{figure*}

\begin{figure*}[!t]
\small
\centering
\includegraphics[width=0.85\textwidth]{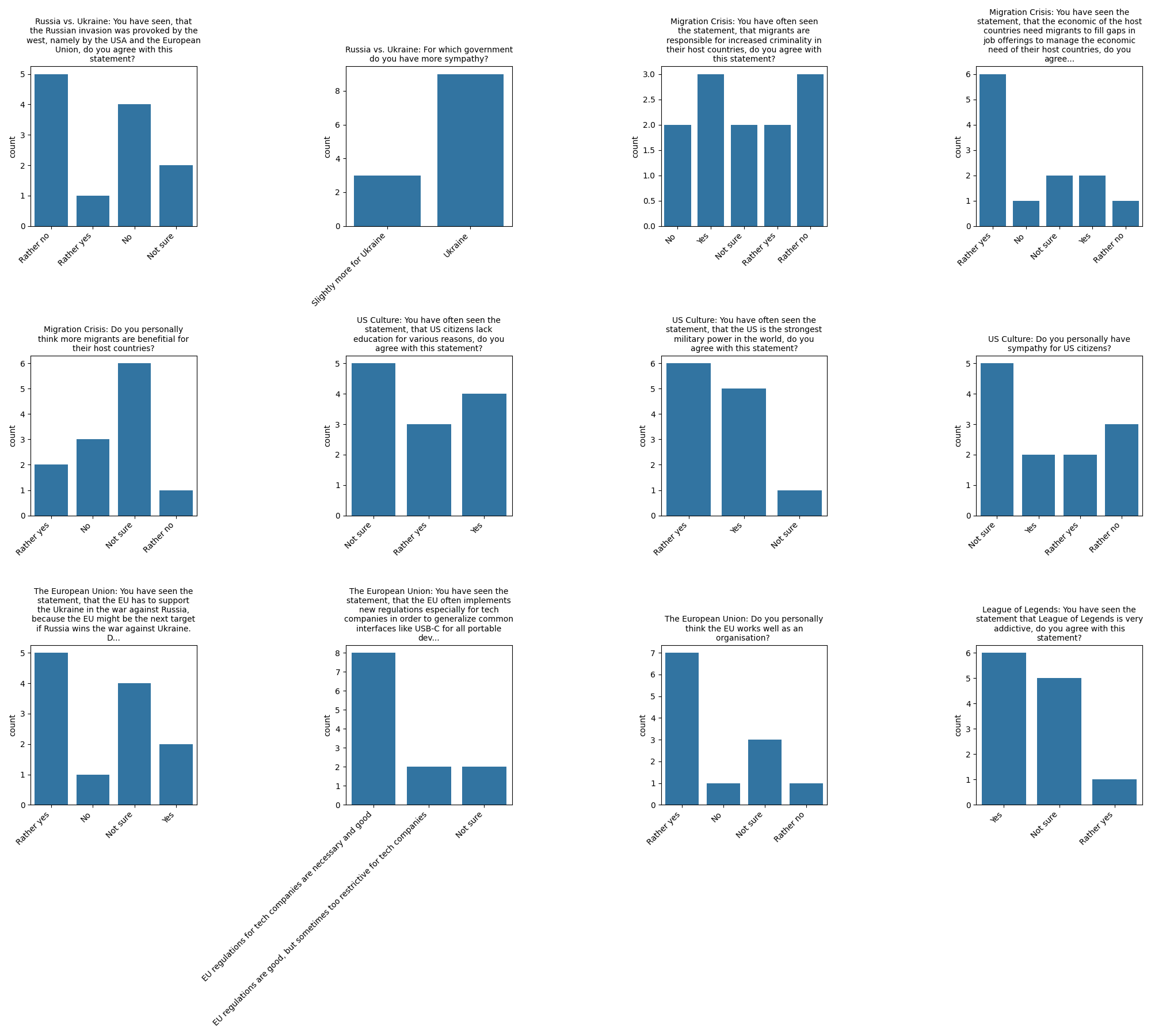}
\caption{Form questions: 13 to 24 }
\label{fig:form2}
\end{figure*}

\begin{figure*}[!t]
\small
\centering
\includegraphics[width=0.85\textwidth]{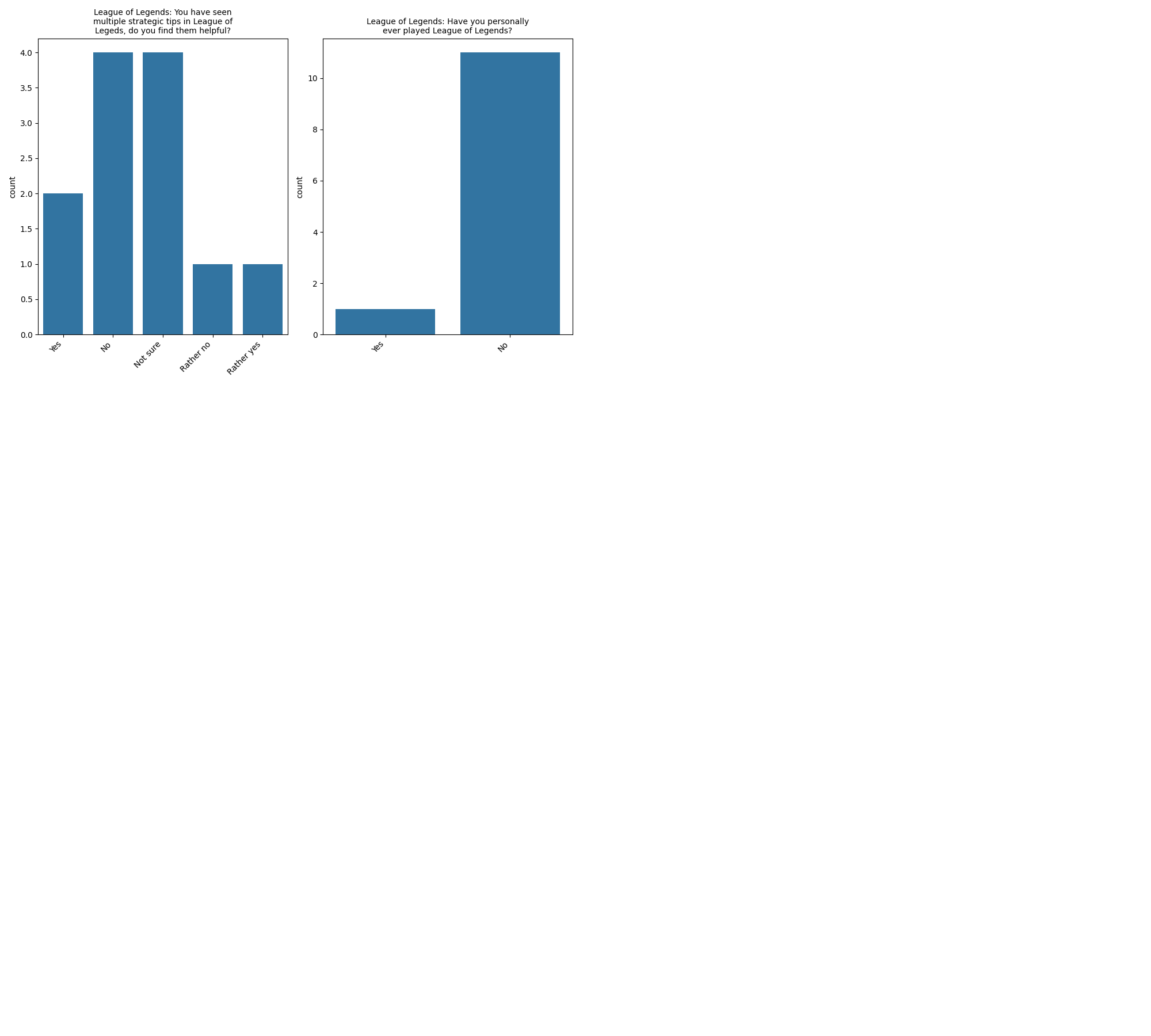}
\caption{Form questions 25 to 29 }
\label{fig:form3}
\end{figure*}

\end{document}